%% file: main.tex
\documentclass{amia}
\usepackage{graphicx}
\usepackage[labelfont=bf]{caption}
\usepackage[superscript,nomove]{cite}
\usepackage{color}
\usepackage{csquotes}
\usepackage[inline]{enumitem}
\usepackage{subcaption}
\usepackage{placeins}
\usepackage{multirow}
\usepackage{amsmath,amsfonts,amssymb}
\usepackage{wrapfig}
\usepackage{floatflt}
\usepackage{url}
\usepackage{setspace}

\input{commands}
\begin{document}

\title{Trajectory Inspection: A Method for Iterative Clinician-Driven Design of Reinforcement Learning Studies}

\author{Christina X. Ji, MEng$^{1*}$, Michael Oberst, MS$^{1*}$, \\Sanjat Kanjilal, MD, MPH$^{2,3}$, David Sontag, PhD$^{1}$}

\institutes{
    $^1$MIT CSAIL, IMES, Cambridge, MA; $^2$Harvard Medical School, Boston, MA; $^3$Harvard Pilgrim Healthcare Institute, Boston, MA;
    $^{*}$ Equal Contribution\footnote{To appear in AMIA 2021 virtual informatics summit}}

\maketitle
\vspace{-6pt}

\noindent{\bf Abstract}
\input{sections/0-abstract}

\input{sections/1-introduction}
\input{sections/2-ml-background}
\input{sections/4-Komorowski-recap}
\input{sections/5-hypothesis-generation}
\input{sections/6-censoring-investigation}
\input{sections/7-discussion}

\subsection*{Acknowledgements}
We would like to thank Monica Agrawal and other members of the Clinical Machine Learning group for valuable feedback on earlier drafts. We would also like to thank the reviewers for their comments. This work was supported by a National Science Foundation CAREER award \#1350965, Office of Naval Research Award No. N00014-17-1-2791, and a KL2 award (an appointed KL2 award) from Harvard Catalyst | The Harvard Clinical and Translational Science Center (National Center for Advancing Translational Sciences, National Institutes of Health Award KL2 TR002542).

\makeatletter
\renewcommand{\@biblabel}[1]{\hfill #1.}
\makeatother

\bibliographystyle{unsrt}

\end{document}

%% file: sections/0-abstract.tex
\textit{Reinforcement learning (RL) has the potential to significantly improve clinical decision making. However, treatment policies learned via RL from observational data are sensitive to subtle choices in study design.  We highlight a simple approach, trajectory inspection, to bring clinicians into an iterative design process for model-based RL studies.  We identify where the model recommends unexpectedly aggressive treatments or expects surprisingly positive outcomes from its recommendations. Then, we examine clinical trajectories simulated with the learned model and policy alongside the actual hospital course.  Applying this approach to recent work on RL for sepsis management, we uncover a model bias towards discharge, a preference for high vasopressor doses that may be linked to small sample sizes, and clinically implausible expectations of discharge without weaning off vasopressors.  We hope that iterations of detecting and addressing the issues unearthed by our method will result in RL policies that inspire more confidence in deployment.}

%% file: sections/1-introduction.tex
\section*{Introduction}

\begin{wrapfigure}[16]{R}{0.5\textwidth}
     \centering
     \vspace{-10pt}
     \includegraphics[width=0.48\textwidth]{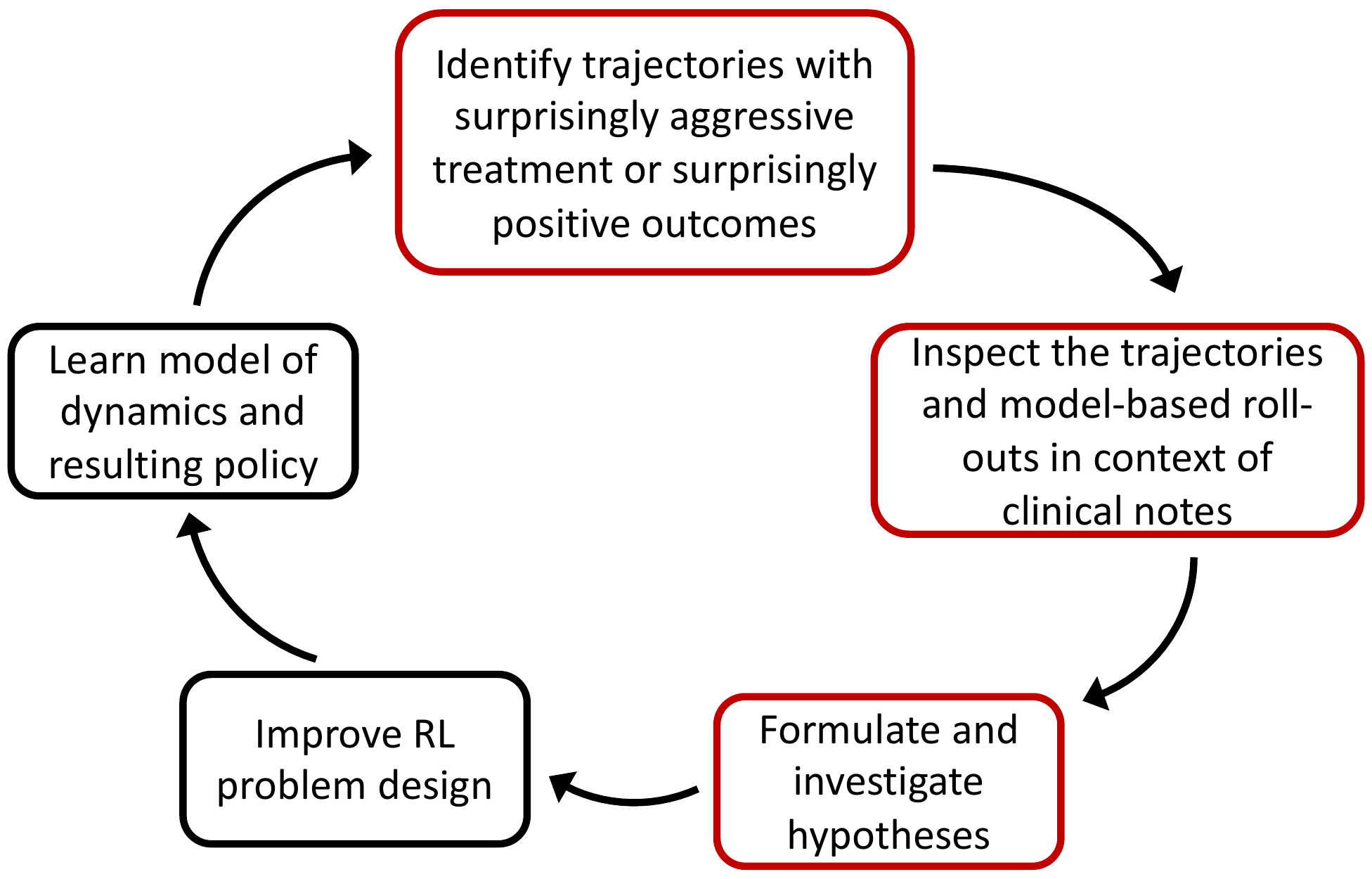}
     \caption{Our proposed method (red boxes) can be integrated into this workflow for improving model-based RL.}
     \label{fig:workflow}
\end{wrapfigure}

Reinforcement learning (RL) has emerged as a popular tool for trying to learn effective policies for managing sequential decisions in patient treatments, including applications in sepsis treatment\cite{Komorowski2018,Raghu2017,Raghu2018}, ventilation weaning\cite{prasad2017reinforcement}, acute hypotension treatment\cite{Futoma2020,Srinivasan2020}, and HIV\cite{parbhoo2017combining}. RL has the potential to provide significant support in clinical decision making. However, in using purely retrospective data, these approaches inherit the usual challenges of learning and evaluating dynamic treatment regimes from observational data, a well-studied topic in epidemiology and bio-statistics\cite{Hernan2020,Chakraborty2013}. For instance, evaluation methods typically assume that all confounding variables are included in the model, and that either existing clinical practice or the dynamics of patient health can be accurately modelled.  Just as in any observational study, if the assumptions do not hold, the analysis can lead to misleading results that could adversely affect clinical practice if adopted. Small effective sample sizes and inadequate specification of the clinical outcome or available actions can further exacerbate the problem\cite{Gottesman2018, Gottesman2019}. Research has been ongoing to overcome these challenges\cite{Futoma2020,Srinivasan2020,futoma2020popcorn}. Because these problems are nuanced, model-checking and interpretability techniques are needed to detect these issues and allow clinicians to actively engage in an iterative process of study design.  We use the term \enquote{study design} to emphasize that this goes beyond choosing a statistical model and encompasses all the design choices involved in translating a clinical decision problem into a RL problem that produces clinically sensible results. By iteratively detecting and fixing these problems until the study is clinically sound, the RL algorithm will be more robust for deployment.

In this work, we highlight a simple approach for examining learned models and policies in model-based RL, a common technique used in recent work to improve clinical decision-making in the management of sepsis\cite{Komorowski2018, Raghu2018}. Our workflow is shown in Figure~\ref{fig:workflow} and proceeds as follows: \begin{enumerate*}[label=(\roman*)]
    \item Select patient cases using one of two heuristics: First, we identify patient states where the learned policy suggests far more aggressive treatment than the observed standard of care, and second, we find patients where the learned policy is predicted to dramatically out-perform the current standard.  We hypothesize that both cases may be due to flaws in the study design.
    \item Contrast the observed trajectories of these patients with the predicted trajectories under the learned model and policy, and flag suspicious model behavior by also comparing predictions against the actual clinical course of patients documented in the medical record.
    \item Investigate questions and hypotheses raised by suspicious model behavior.
\end{enumerate*} The code for reproducing our work is located at \url{https://github.com/clinicalml/trajectory-inspection}.

We demonstrate the utility of this approach by applying it to recent work published in \textit{Nature Medicine} \cite{Komorowski2018} that we replicate using MIMIC-III, a freely available ICU dataset\cite{Johnson2016}. We generalize our anecdotal observations through aggregate analysis and discover that the data pre-processing heuristic used in recent works on RL for sepsis management\cite{Komorowski2018, Raghu2017} can lead to biases in estimating the model.  We also observe that high recommended vasopressor doses may be linked to small sample sizes, and that the model has clinically implausible expectations, such as patient discharge from the ICU within 4 hours of receiving large doses of vasopressors.

%% file: sections/2-ml-background.tex
\section*{Background}
\textbf{Reinforcement learning methods}: The approach that we replicate\cite{Komorowski2018} learns to manage sepsis using RL\cite{kaelbling1996,sutton2018}.  The framework is a \textit{Markov decision process (MDP)}.  This model specifies a set of possible states and actions called the \textit{state space} and \textit{action space}, respectively. The state contains patient features, such as demographics, vital signs, and prior treatment. The action corresponds to a treatment decision. At each time step, an action is taken based on the current state, and the patient transitions to a new state. This transition model captures how the patient responds to treatment and evolves over time. The \textit{Markov property} states that this transition depends only on the most recent state and action and is otherwise independent of history. The \textit{reward function} specifies how each transition maps to a positive or negative outcome.  Each sequence of states and actions and the reward forms a \textit{trajectory} and represents the course of the patient stay. A \textit{policy} specifies which action to take from each state. This can be a \textit{behavior policy} observed in the training samples or a more optimal policy.  The goal is to learn a policy that maximizes the reward. 

\begin{figure}[t]
     \centering
     \includegraphics[width=0.88\textwidth]{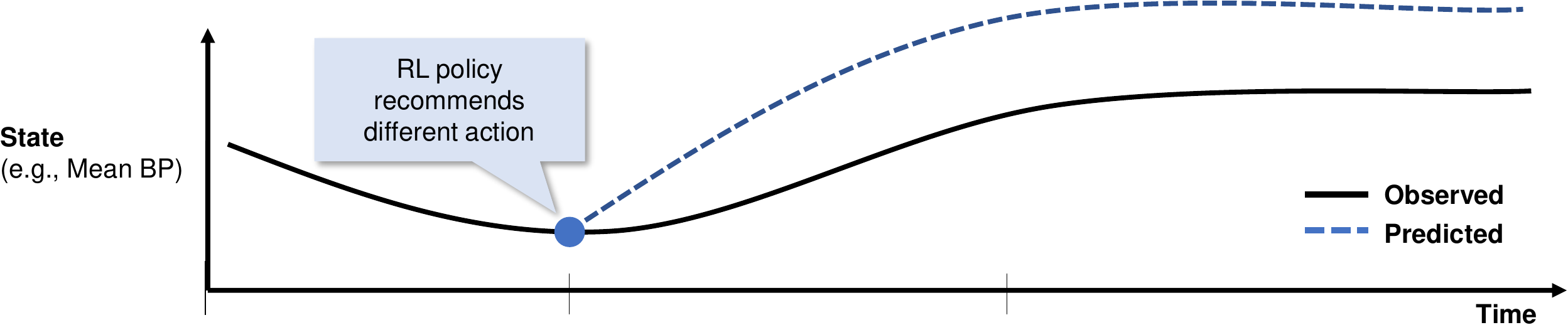}
     \caption{Conceptual illustration of a patient trajectory (in black) and a model-based roll-out (in blue) which tracks what the RL model predicts would occur as a result of the actions considered `optimal' by the learned policy.}
     \label{fig:mb_example}
\end{figure}

\textbf{Model-based trajectories}: In \textit{model-based} RL, a transition model and reward function are estimated, and the policy is learned using those estimates.\footnote{Alternatively, \textit{model-free} RL still relies on modelling existing practice, but no explicit transition model is learned.} We can generate \textit{model-based trajectories} (sometimes referred to as \enquote{roll-outs}) given an initial state, a transition model, and a policy. These trajectories are predictions, made by the model, for how a patient would progress under the policy.  Concretely, at each time step, an action is drawn from the distribution specified by the policy for the current state. Then, the next state is drawn from the distribution specified by the transition model for that state-action pair. This process repeats until an absorbing state (with a reward) or a maximum length of 20 steps is reached. Because the transitions and sometimes the policy are probabilistic, multiple different model-based roll-outs can be generated from the same initial state. A conceptual illustration is shown in Figure \ref{fig:mb_example}. 

\textbf{Notable challenges with off-policy evaluation}: Just as with any observational study, confounding and other biases can lead to unreliable estimates of the value of a policy.  As a conceptual illustration, imagine that all patients are either \enquote{healthy} or \enquote{sick}, that doctors tend to aggressively treat sick patients, and that sick patients have higher mortality rates than healthy patients, even with aggressive treatment.  In this setting, if we do not adjust for comorbidities and severity of the presenting illness (by including it in the state space of our model), then our evaluation might wrongly conclude that \enquote{never treating} patients is a good policy, due to the association between treatment and mortality.

Indeed, this phenomenon has been observed in the literature on applying RL to sepsis management.  In some cases, it has been noted that a \enquote{zero-drug} policy has a high estimated value relative to current practice, using the same evaluation methodologies that are used to justify the value of RL policies\cite{Komorowski2018,Jeter2019,Komorowski2019}.  In other instances, the RL policy associates the more intensive treatments observed for high acuity patients with high mortality and instead recommends less intensive treatments that are rarely observed\cite{Gottesman2018}.  Possible unmeasured confounding is not the only hurdle to performing valid inference. Other challenges include small effective sample sizes and inappropriate reward definitions\cite{Gottesman2019}.

\textbf{Related work}: We differentiate our approach from recent work in counterfactual off-policy evaluation\cite{Oberst2019}. While that work also involves simulating from a model and has a similar heuristic for selecting trajectories, it requires additional causal modelling assumptions, and we view our approach as simpler to apply.  More importantly, we demonstrate our approach on a real dataset and model, while that work only considered a synthetic dataset.

%% file: sections/4-Komorowski-recap.tex
\section*{Methods: Setup}
\textbf{Main outcome measurement}: Our primary objective is necessarily qualitative, demonstrating that our approach can help researchers better understand the challenges in designing RL studies and give them a concrete tool to integrate clinician input into improvement of study design. That said, we do provide quantitative metrics related to our trajectory selection heuristics, e.g., what it means for a recommended action to be surprisingly aggressive, or a model-based roll-out to have a surprisingly positive outcome (see Figure~\ref{fig:heuristic_overviews}).  We also compute several diagnostic metrics during our investigation of specific questions raised by trajectory inspection.

\textbf{Design/patient cohort}: Our work is a secondary use of electronic health record data, specifically the MIMIC-III dataset\cite{Johnson2016}. As we are replicating Komorowski et al (2018)\cite{Komorowski2018}, we use their code\footnote{\url{https://github.com/matthieukomorowski/AI_Clinician}} for defining the sepsis cohort: adults fulfilling the sepsis-3 criteria\cite{singer2016third,seymour2016assessment}, including antibiotic prescription, lab work for signs of infection, and a sequential organ failure assessment (SOFA) score $\ge$ 2. Patients with missing fluids or mortality data are excluded. Patients who satisfy all of the following conditions are also excluded as they may have been placed on comfort measures: (a) died within 24 hours of the end of the data collection window, (b) received vasopressors at any point, and (c) whose vasopressors were stopped at the end of the data collection. Our cohort consists of 20,090 ICU admissions. 

\textbf{Setup of Komorowski et al (2018)\cite{Komorowski2018}}
\label{sec:komorowski_setup} The state space includes demographics (e.g., age), vital signs (e.g., blood pressures, sequential organ failure assessment), lab values (e.g., white blood cell count, creatinine), ventilation parameters (e.g., FiO2), and treatment information (e.g., fluid balance and output aggregated over 4-hour intervals). The policy takes in these state features and gives a recommendation on vasopressors and fluids. We follow the publicly available code provided by the authors and use the same version of the MIMIC-III dataset\cite{Johnson2016} in our replication. In this section, we review some details of the set-up, focusing on modelling decisions. Other details can be found in the original paper.

Trajectories are discretized into 4-hour intervals and limited to a maximum of 20 time steps, representing up to 28 hours before and 52 hours after onset of sepsis\footnote{The paper reports using 24 hours before and 48 hours after. Our implementation follows their publicly released code, which uses 28 and 52.}. Starting from a set of continuous and discrete variables\footnote{Our replication follows the publicly available code, which uses 47 (excluding Elixhauser Index) state variables instead of the stated 48.}, the state space is discretized using k-means clustering with 750 clusters, and two absorbing states are added for 90-day survival and 90-day mortality (including in-hospital mortality).  Model actions are limited to providing blood pressure support through intravenous fluids or adjusting vasopressors. The action space contains 25 discrete choices, corresponding to either zero or one of four quartiles of total fluid input and maximum vasopressor dosage over each 4-hour time step. 
Rewards are defined as $\pm 100$ depending on 90-day mortality and are recorded in the retrospective trajectories as a transition into an absorbing state described above.  After replicating this work, we have a transition model, a behavior policy, and a target policy, learned from observational data.

%% file: sections/5-hypothesis-generation.tex
\section*{Methods: Selecting trajectories for inspection} As previously described, our approach relies on inspecting clinical trajectories alongside their model-based roll-outs. We propose two heuristics for selecting trajectories: (i) \textit{Surprisingly aggressive treatments}: We find states where the RL-recommended action is rarely observed and more aggressive than the observed practice of clinicians. We then start the roll-outs where those states occur in trajectories. (ii) \textit{Surprisingly positive outcomes}: We select clinical trajectories with initial states that have the largest differences between predicted and actual outcomes.

\begin{figure}[t]
\centering
    \begin{subfigure}{0.4\linewidth}
    \centering
    \includegraphics[width=0.96\linewidth,trim=0 20 0 0,clip]{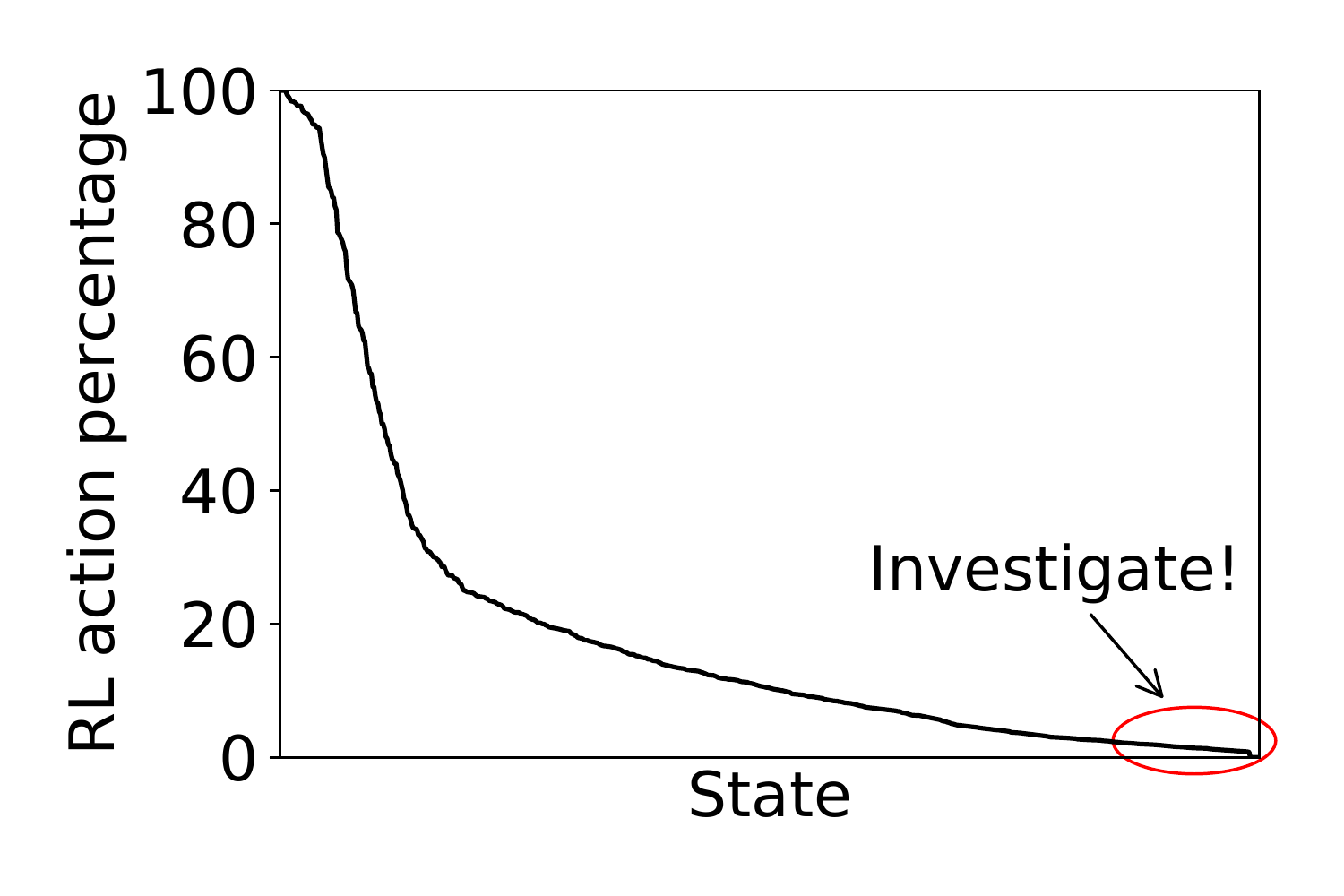}
    \caption{}
    \label{fig:rl_action_percentages}
    \end{subfigure}~%
  \begin{subfigure}{0.56\linewidth}
      \centering
      \includegraphics[width=0.96\linewidth,trim=0 20 0 0,clip]{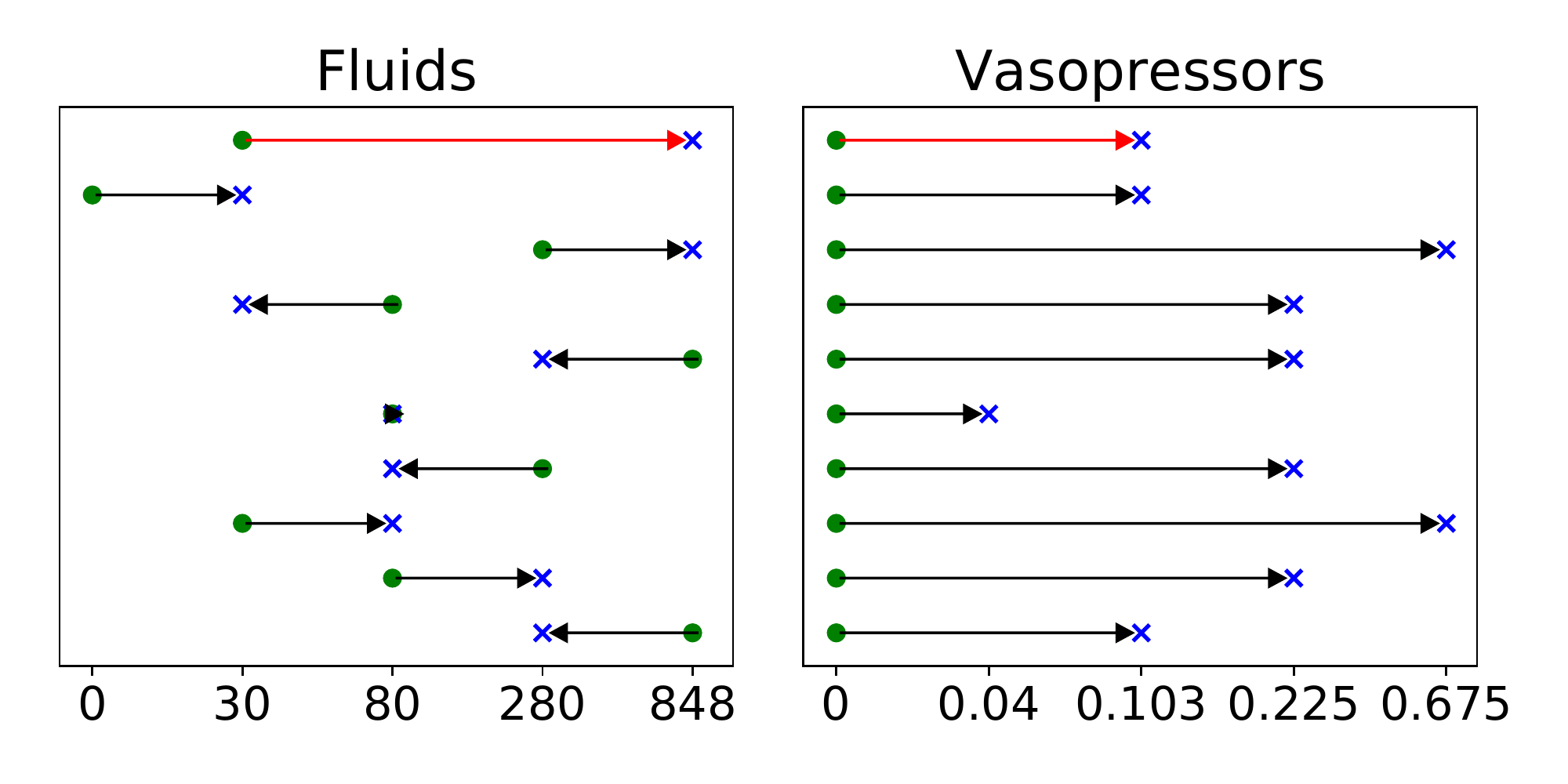}
      \caption{}
      \label{fig:heuristic1}
  \end{subfigure}
  \begin{subfigure}{0.9\linewidth}
      \vspace{2em}
      \centering
      \includegraphics[width=0.6\linewidth]{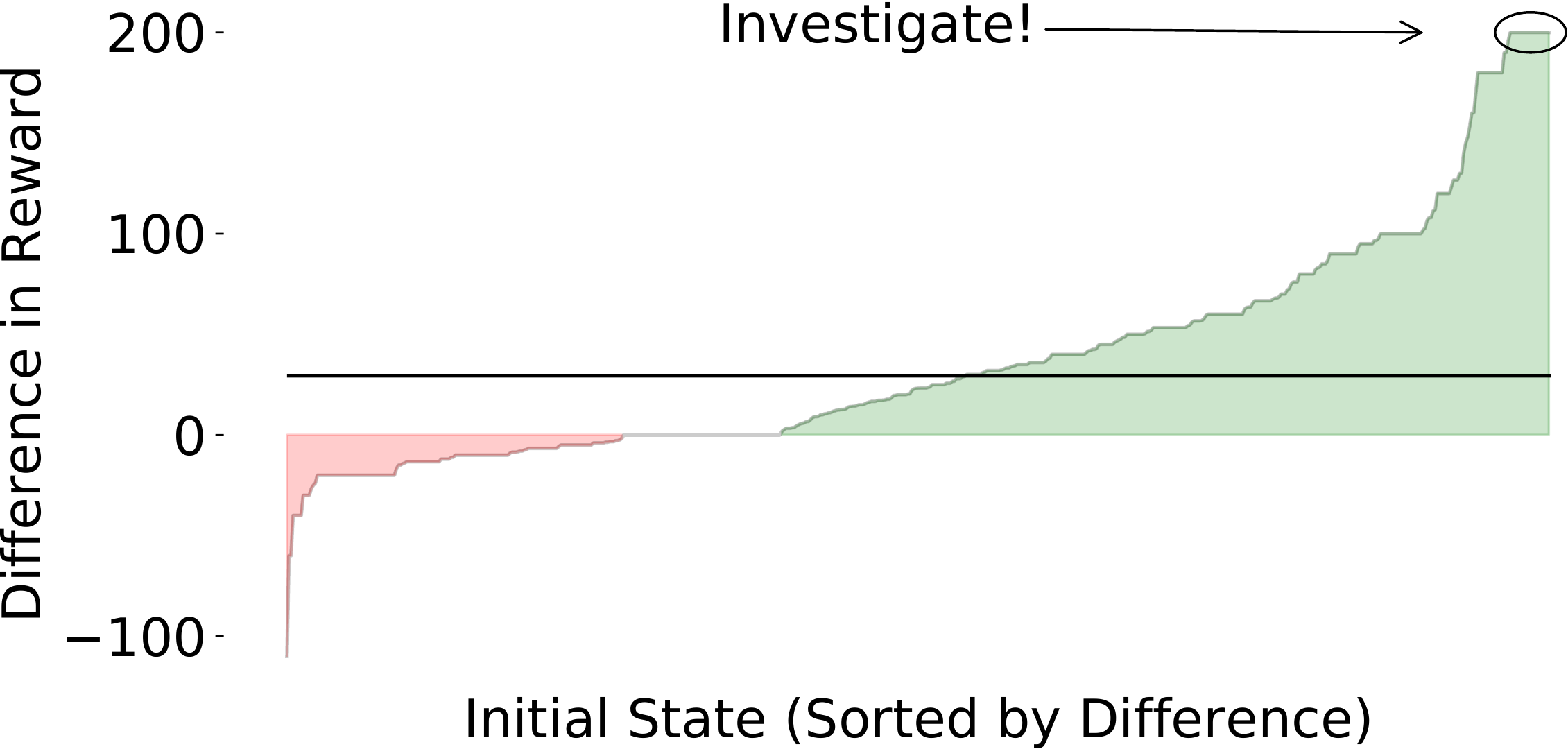}
      \caption{}
      \label{fig:heuristic2}
  \end{subfigure}
  \caption{Visualization of our heuristics for trajectory selection. (a) Observed frequency of the RL action for each state in the training data. States are sorted by this quantity on the X-axis. (b) Differences between the RL action (blue cross) and the action most commonly observed in the training data (green dot) from the 10 states whose RL action was rarest. The X-axis is total milliliters over the past 4 hours for fluids and micrograms per kilogram of body weight per minute for vasopressors. Each row is a state, with the one we inspect in red. (c) Differences in predicted outcomes.  X-axis sorts states by the average difference between model-based and observed rewards for trajectories with that initial state. The black line denotes the overall average difference weighted by initial state.}
  \label{fig:heuristic_overviews}
\end{figure}

\textbf{Selecting states with surprisingly aggressive treatments}: To select trajectories, we identify instances where the learned policy recommends aggressive treatment even though conservative management is observed. Concretely, across all states, we first look for actions recommended by the learned policy that occur at most 1\% of the time in the training data at that state, as visualized in Figure~\ref{fig:rl_action_percentages}.  This yields 15 states, and we plot the differences between the most common clinician action and the RL action of the top 10 in Figure~\ref{fig:heuristic1}. We analyze a state where the patient receives significantly higher amounts of fluids and vasopressors under the RL policy as compared to standard clinical practice. The clinical parameters that characterize this state are mostly within normal limits except for a mildly elevated white blood cell count. The most common action performed by the clinicians for patients in this state was to provide 30cc of fluids, which is suggestive of a very low infusion necessary to maintain the patency of the IV line, and no vasopressors. In contrast, the RL policy recommends 848cc of fluids and 0.13 micrograms/kg/min of vasopressors. Of the 632 times the state is observed in the training data, the common action noted above is taken 161 times, while the `optimal' action according to the learned RL policy is chosen only 6 times by clinicians. For these trajectories, we start the model-based roll-outs at the time step where the selected state and common action occurred.

\textbf{Selecting initial states with surprisingly positive outcomes}: To select trajectories under this heuristic we identify patients where the model expects that the new policy will most out-perform the observed current practice.  We hypothesize that by focusing on extremes, we will enrich for instances that highlight problems with the RL model. Concretely, we create model-based roll-outs for each actual trajectory in the test set, starting from the observed initial state.  We then select trajectories with substantially higher model-based reward than the actual reward.
In particular, for each state, we examine all trajectories that start in that state and take the mean difference between (a) the reward of 5 model-based roll-outs and (b) the observed reward. We then inspect trajectories for which this difference is largest. In Figure~\ref{fig:heuristic2}, we visualize this difference across individual states and highlight some initial states for inspection. We choose one of the highlighted initial states, where 90-day mortality is observed in the actual data, but all model-based roll-outs (from the same initial state) end in 90-day survival.  We investigate a clinical trajectory with this initial state.

\section*{Results: Examining trajectories alongside the medical record}

In this section, we present two patient trajectories, selected using the heuristics described above.  In both cases, we examine the full set of clinical notes available for the patient stay, along with their actual trajectories of vital signs. We compare this against the model-based trajectories predicted by the RL model under the learned RL policy. We can ask \textit{whether these model-based predictions seem clinically sensible}, particularly in light of the notes.

\begin{figure}[t]
\begin{subfigure}{0.54\linewidth}
\begin{quote}
\small
    Transferred in from outside hospital after undergoing cardiac catheterization that revealed coronary artery disease. She \textelp{} was brought to the operating room for {\bf coronary artery bypass graft surgery}. \textelp{} She was {\bf transferred to the intensive care unit for post operative management}.  In the first twenty four hours she was weaned from sedation, awoke neurologically intact, and was {\bf extubated without complications. She was started on betablockers and gently diuresed toward preoperative weight. On post operative day one she was transferred to the floor.} Chest tubes and pacing wires were discontinued without complication. \textelp{} By the time of discharge on POD 5 the patient was ambulating freely, the wound was healing and pain was controlled with oral analgesics. {\bf The patient was discharged to home in good condition}
    \end{quote}%
    \vspace{-10pt}
    \caption{}
    \label{fig:snippets2}
\end{subfigure}
\hspace{-30pt}
\begin{subfigure}{0.51\linewidth}
\begin{quote}
\small
    Pt had Septic physiology on admission \textelp{} and the most likely source was felt to be postobstructive pneumonia and parapneumonic effusion \textelp{} Pt was intubated \textelp{and} R-sided chest tube initially drained a large volume of mucinous fluid \textelp{} noncontrast chest CT showed a RLL rounded opacity \textelp{} CT chest with contrast showed the RLL opacity was a large mass with significant surrounding lymphadenopathy and marked pleural tumor, with possible invasion into the chest wall \textelp{} {\bf pt was ultimately diagnosed with stage IIIA lung cancer.}
    Pt's progressive respiratory failure was treated with BiPAP and nebs
    \textelp{} {\bf Pt ultimately died of respiratory failure likely due to a combination of COPD, pneumonia and lung cancer}
    \end{quote}
    \caption{}
    \label{fig:snippets_1}
\end{subfigure}
\caption{Selected extracts from the de-identified medical record for Patient 1 (a) and Patient 2 (b).  Emphasis added.  Acronyms: POD, post-operative day. Pt,  patient. RLL, right lower lobe. BiPAP, Bilevel Positive Airway Pressure}
\label{fig:snippets_both_patients}
\end{figure}

\begin{figure}[t]
  \centering
  \begin{subfigure}{0.495\linewidth}
  \includegraphics[width=\linewidth, trim=0 20 550 46,clip]{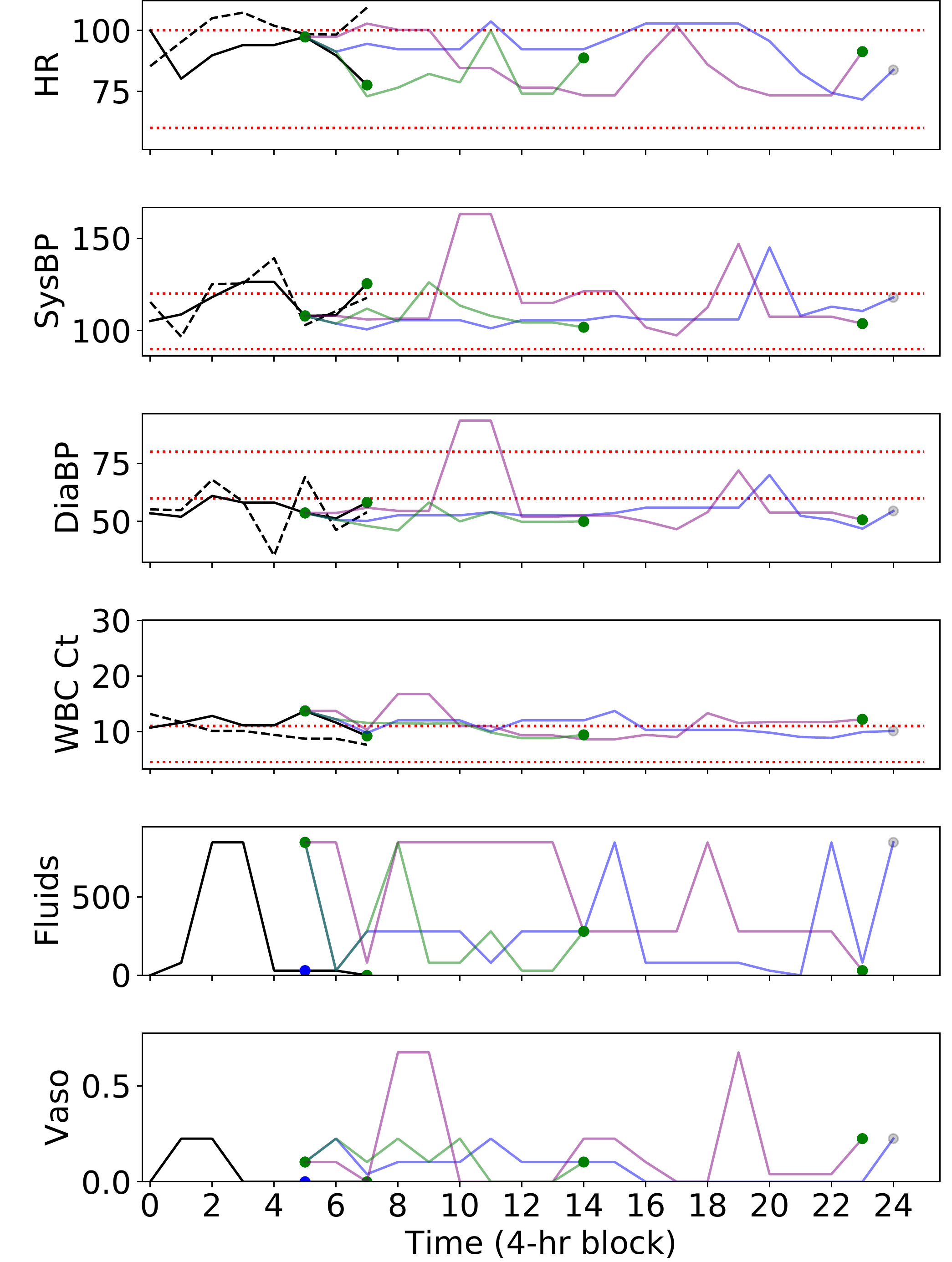}
  \caption{}
  \label{fig:trajectory_1}
  \end{subfigure}~%
  \begin{subfigure}{0.45\linewidth}
  \includegraphics[width=\linewidth, trim=605 20 0 46,clip]{figs/two_patients_final.pdf}
  \caption{}
  \label{fig:trajectory_2}
  \end{subfigure}
  \caption{Comparison of actual trajectories and model-based roll-outs for patient 1 (a) and patient 2 (b).  Five model-based trajectories per patient start at the blue dots:  In (a), two of these end when they start at time step 5, and in (b), two end at time step 0.  Model-based roll-outs are in various colors, and vitals are derived from the k-means medians. At the ends of the trajectories, green dots indicate 90-day survival, red crosses indicate 90-day mortality, and grey dots indicate maximum allowed length without discharge. Black dotted lines show actual values, while black solid lines denote median values from k-means clustering. Red dotted lines are reference ranges. HR: heart rate. SysBP: systolic blood pressure. DiaBP: diastolic blood pressure. WBC Ct: white blood cell count. Vaso: vasopressors. Fluids dosage in total milliliters over the past 4 hours and vasopressor levels in micrograms per kilogram of body weight per minute.}
  \label{fig:both_patient_trajectories}
\end{figure}

\textbf{Clinical review of patient 1: Surprisingly aggressive treatment}:  We present a snippet from the de-identified medical record in Figure~\ref{fig:snippets2}, and in Figure~\ref{fig:trajectory_1}, we present selected vital signs, along with the actions.  We have the following takeaways from the notes: 
(i) \textit{Cause of admission:} This patient presented with shortness of breath and chest pain after a previous visit had revealed coronary artery disease.  As a result, she underwent coronary artery bypass graft (CABG) surgery and was transferred to the ICU for post-operative management on the same day as the surgery.  
(ii) \textit{Treatment during ICU stay}: Around 12 hours into the ICU stay, radiology notes indicated signs of pulmonary edema. After 24 hours, the patient was recovering well, weaned from sedation, and extubated without complications. The patient received cardio-protective beta blockers on post-operative day 1, suggesting that she was in a stable condition from a hemodynamic standpoint. Stability was further evidenced by the fact that diuretics were used to gently remove fluids and bring her volume status back to pre-surgery levels.  The patient was transferred out of the ICU and sent home without pulmonary edema and in good condition on post-operative day five.

In light of these points, we make the following observations regarding the trajectories in Figure \ref{fig:trajectory_1}: 
(i) \textit{The RL policy suggests unnecessary prolonged use of aggressive treatments.} In the actual trajectory, clinicians initiated IV fluids and vasopressors approximately 8 hours into the ICU stay and in the immediate post-operative period. By the time we start model-based roll-outs, 20 hours into the ICU stay, the amount of fluids and vasopressors had already been greatly reduced or completely discontinued given the patient's uncomplicated recovery.\footnote{All the time steps are 4 hours apart, except there are 20 hours between steps 6 and 7.} Despite the benign hospital course for the patient, the RL policy recommended restarting larger dosages of both fluids and vasopressors, which would likely have increased risk of pulmonary edema and fluid overload. 
(ii) \textit{Expected patient discharge while on vasopressors is clinically implausible}: In two of the model-based trajectories, the model anticipates that the patient will leave the ICU within 4 hours of giving vasopressors at the start of the roll-outs, and in most of the model-based trajectories, the patient is discharged while on vasopressors. This goes against clinical intuition that a patient on vasopressors should be weaned off and monitored prior to leaving the ICU.
(iii) \textit{Response in state variables is inconsistent with action.} The effect of vasopressors becomes apparent within half an hour, as evidenced by how blood pressure rose at time step 2 following a dosage at the previous time step in the actual trajectory. In the model-based trajectories, however, vasopressors are administered at time step 5, but all of the trajectories show little change in blood pressure at the next time step. For the model-based roll-out indicated in purple, 0.675 micrograms of vasopressors per kilogram of body weight per minute were administered at 32-36 hours, which is even higher than the doses for the other model-based trajectories. 
However, blood pressure does not rise until 40-44 hours. This may indicate that the model is not accurately modelling the drug response. The rationale for the RL policy is also unclear given the lack of a clear indication for blood pressure support.

\textbf{Clinical review of patient 2: Surprisingly positive outcomes}: We present a selected snippet from the medical record in Figure~\ref{fig:snippets_1}, and in Figure~\ref{fig:trajectory_2}, we present selected vital signs, along with the actions.  We summarize the major takeaways here from the full medical record:
(i) \textit{Cause of admission:} This patient was admitted after collapsing, thought secondary to either respiratory or cardiac failure. The patient was taken immediately to the cardiac catheterization lab\footnote{A cath lab is an exam room with diagnostic imaging equipment to visualize the heart.}, where a myocardial infarction due to coronary artery disease was ruled out.  Chest imaging showed a large amount of fluid around the right lung and a large mass in the lower right lobe.  This was later discovered to be Stage IIIA lung cancer\footnote{Stage IIIA lung cancer is defined as spread to nearby lymph nodes, but not other organs.}. The sum of the diagnostic studies suggested that the most likely etiology of the patient's presentation was cardiovascular collapse\footnote{Cardiovascular collapse is the rapid or sudden development of cardiac failure. This was a symptom that resulted from the tumor.} and a secondary post-obstructive pneumonia that were both due to the mass effect of the tumor.
(ii) \textit{Treatment before and during ICU:} The pleural effusion was felt to be multifactorial and likely due to poor forward flow as well as inflammation from the adjacent pneumonia\footnote{The pleural space is the thin fluid-filled space between the two membranes around each lung}. A chest tube was placed, which subsequently drained $>$1L of exudative serous fluid. The patient's clinical status responded rapidly, suggesting the external compression from the fluid was a major contributor to his course.  Upon transfer to the ICU, physicians continued to administer vasopressors and antibiotics, with the former being gradually weaned starting 44 hours into the trajectory. 
(iii) \textit{Cause of death:} Despite the placement of a chest tube, the underlying problem of a large lung mass leading to cardiovascular compromise remained unaddressed. In part due to the morbidity of the necessary chemotherapy, the providers, the patient, and the family decided that further aggressive interventions would not have been in the patient's interests and he was made `comfort measures only' 12 days after the end of this trajectory. He passed away shortly thereafter.

In light of these points, we make the following observations regarding the visualization in Figure~\ref{fig:trajectory_2}: \begin{enumerate*}[label=(\roman*)]
    \item \textit{Surprisingly early termination of several trajectories:} Two of the model-based trajectories end after the first 4 hours, and another ends after the first 12 hours. However, the average length of an ICU stay is 3.3 days\cite{Hunter2014}.
    \item \textit{The anticipated outcomes are not credible given the medical record}: The cause of death in this patient was irreversible lung damage caused by Stage IIIA lung cancer and pneumonia. As such, the only interventions that would have resulted in survival would have been aggressive chemotherapy, careful cardiovascular support, and a short course of antibiotics. Restrictive fluid and vasopressor therapy (the RL policy) on its own would be very unlikely to have a major salutary effect on the clinical course. Yet, all model-based trajectories result in subsequent 90-day survival.
\end{enumerate*}

%% file: sections/6-censoring-investigation.tex
\section*{Results: Investigation of questions raised by trajectory inspection}
\label{sec:deep_dive}

Although our anecdotal analysis from investigating specific trajectories does not prove that something is fundamentally amiss with the model, the examples serve as inspiration for improving study design. We examine the following three questions suggested by the previous section: 
(i) \textit{Why do trajectories seem to end early?} For patient 1, we observed that two of the five model-based roll-outs ended within 4 hours.  Similarly for patient 2, we observed several trajectories that ended surprisingly early.  This may suggest that the model is overly optimistic about how quickly patients will be discharged.
(ii) \textit{Why does the model learn to use uncommonly high vasopressor doses?}
The RL-recommended aggressive treatment of patient 1 did not appear to have a clinical basis. Indeed, while selecting trajectories via our \enquote{surprising treatment} heuristic, we observed in Figure~\ref{fig:heuristic1} that these RL-recommended actions tend to involve higher levels of vasopressor support. 
(iii) \textit{Why does the model learn to expect discharge while on vasopressors?} In several model-based roll-outs for patient 1, the model predicts discharge (and positive outcomes) to occur while the patient was still on vasopressors.
In the remainder of this section, we move from these anecdotal observations to a more thorough analysis across the entire dataset and seek to understand the answers to these questions. 

\begin{figure}[t]
  \centering
  \includegraphics[width=0.8\linewidth]{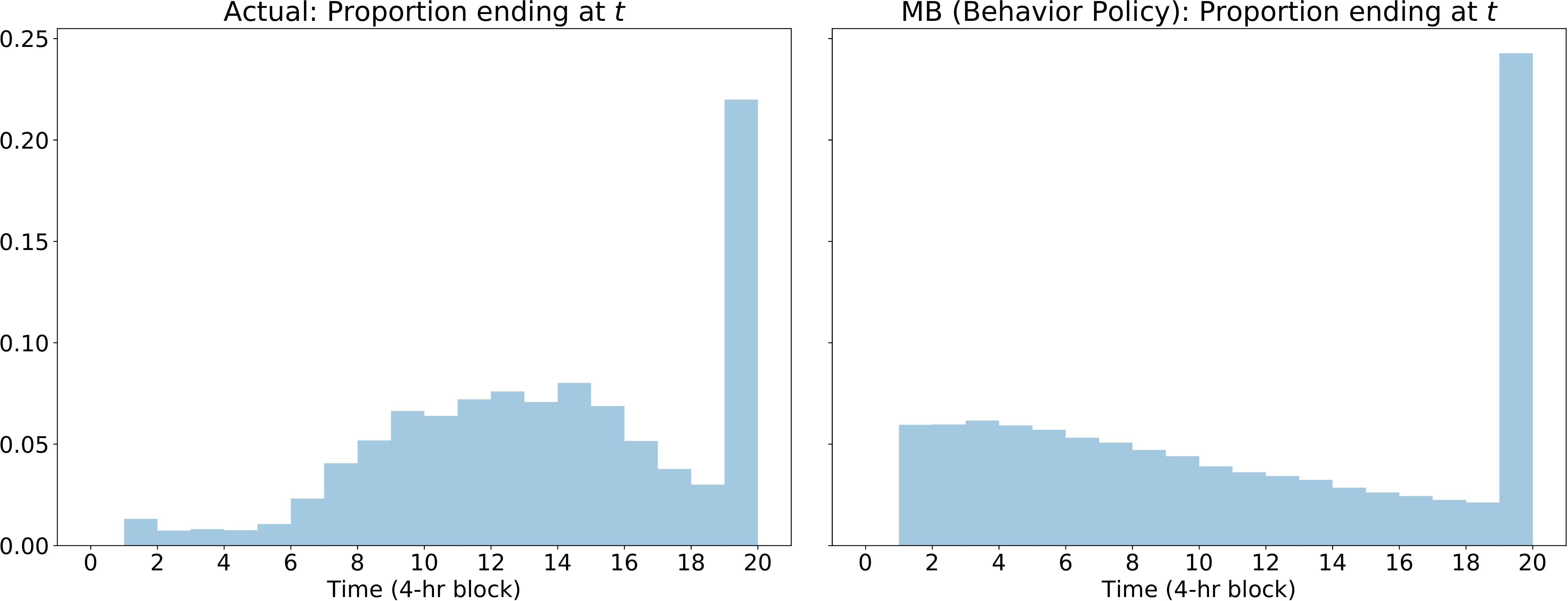}
  \caption{Histogram of trajectory lengths, comparing the training data on the left with the model-based roll-outs under the behavior policy on the right from the same distribution of initial states.}
  %Time indicates 4-hour blocs, so e.g., $t = 2$ indicates 8 hours.}
  \label{fig:traj_lengths_actual_vs_mb_train}
\end{figure}

\textbf{The model is biased towards early discharge, in part due to censoring}: We seek to demonstrate that early predicted discharge is common and due to the transition model rather than the RL policy. To do so, we compare the lengths of training trajectories and model-based roll-outs under the \textit{behavior policy} (the actions taken by clinicians) in Figure~\ref{fig:traj_lengths_actual_vs_mb_train}.  Because the transition model and behavior policy are derived from the training data, we should expect close matches in the distribution of trajectory lengths if the model is a good fit to the data.  Instead, we make the following observations: First, a large proportion of trajectories in the training data are \enquote{censored}, i.e., they do not end within the defined time window.
Second, shorter trajectories are indeed more common in the model-based roll-outs, where the distribution of lengths resembles a regular decay rather than the \enquote{bell shape} observed in the training data.\footnote{While clear when examining the distribution of trajectory lengths, this model mismatch may not be as obvious if one only looks at average length of trajectories, as done in the \enquote{Goodness of fit of the transition matrix} analysis in follow-up work\cite{Komorowski2019} to the model we replicate} 

\begin{figure}[t]
  \centering
  \begin{subfigure}{0.45\linewidth}
  \includegraphics[width=\linewidth]{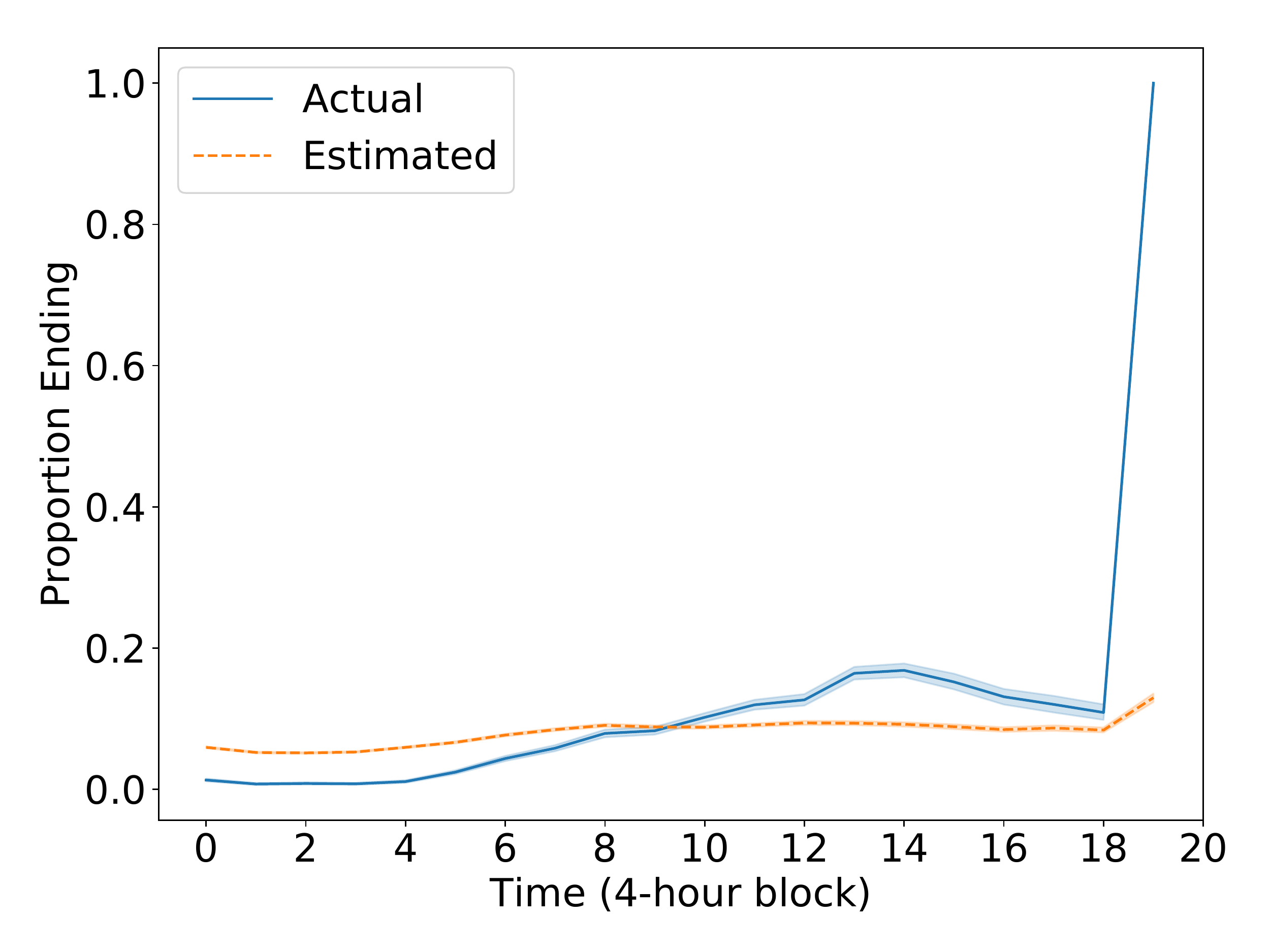}
  \caption{}
  \label{fig:prop_end1}
  \end{subfigure}~%
  \begin{subfigure}{0.45\linewidth}
  \includegraphics[width=\linewidth]{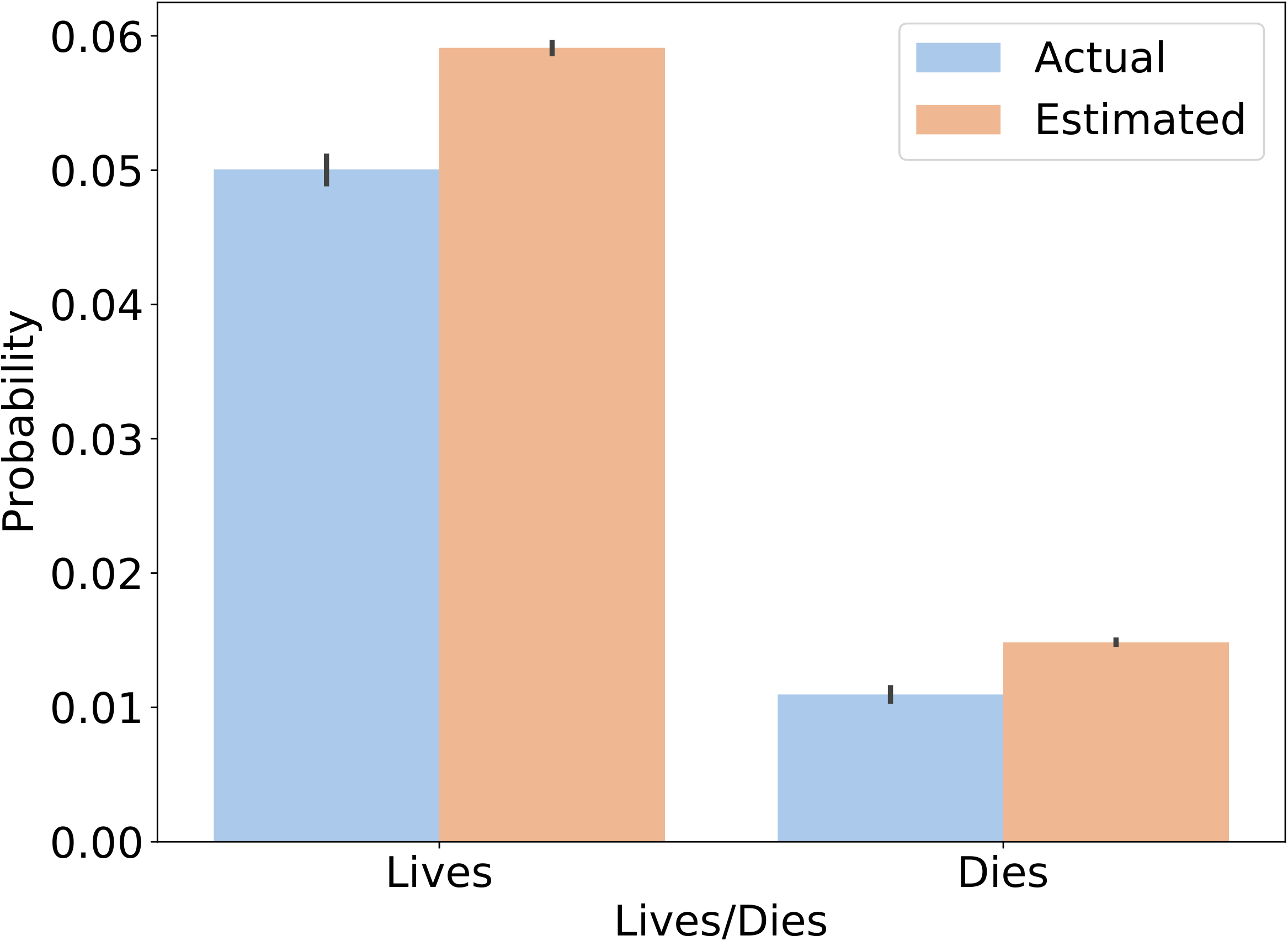}
  \caption{}
  \label{fig:prop_lives_dies_bars}
  \end{subfigure}
  \caption{(a) At each time point, the actual proportion of remaining trajectories that immediately terminate (in blue) vs the average predicted probability of immediate termination (in orange).  The latter is calculated using the learned transition model and the states and actions that are observed at that time step. (b) Across all time points before the 20th step, we compute the actual and estimated average probabilities of immediate termination.  For both (a) and (b), 95\% confidence intervals are generated with 1000 bootstrapped samples using the seaborn package in Python.}
  \label{fig:prop_end}
 \end{figure}

Third, we argue that early termination of model-based roll-outs arises from the heuristic combination of (i) censoring of trajectories after 20 time steps and (ii) the addition of a pseudo-transition at the end of each censored trajectory to either the 90-day survival or mortality absorbing state.  We refer to this heuristic as \textbf{censoring with terminal rewards}. To demonstrate that this heuristic leads to a bias towards discharge in the learned transition model, we focus on the transition to an absorbing state (corresponding to end of trajectory with 90-day survival or 90-day mortality). Figure~\ref{fig:prop_end1} compares the mean estimated probability of trajectory termination to the observed proportion at each time step.  We note in particular that the model cannot capture the fact that at the final time step \textit{every remaining trajectory will end}, due to the censoring with terminal rewards heuristic. To match the overall termination probability it observes, the model is forced to systematically over-estimate the probability of transition to an absorbing state in the time steps prior to the 20th one as seen in Figure~\ref{fig:prop_lives_dies_bars}. If we interpret the transition to an absorbing state as the event \enquote{patient is discharged from the ICU and then lives/dies in the subsequent 90 days}, then most observed \enquote{discharges} at the 20th step do not reflect an actual discharge from the ICU, and this can mislead the model. % To understand possible clinical implications, consider a conceptual example:  
For instance, suppose that many patients are in critical condition at the 20th step and are receiving aggressive interventions. In reality, these trajectories likely continue as the physicians work to stabilize these patients.  However, because the processed data implies that these patients are immediately discharged, this can lead the model to believe that when another patient is in critical condition, even at the start of a trajectory, there is some moderate probability that their stay will immediately end.

\textbf{High vasopressor doses may be linked to small sample sizes}: The RL model does not consider sample size. In the work that we replicate\cite{Komorowski2018}, actions that are taken less than 5 times from a particular state are removed from the training data.\footnote{If all actions at a state occur less than 5 times, the RL policy is set to give no treatment. We omit such states from our analysis in this paragraph.} Among the training data, the RL action from the state analyzed for patient 1 was observed for only 6 patients (who survived), while the common action was observed for 148 patients who survived and 13 patients who died. Thus, the model learns that the rare RL action leads to better outcomes on average. This selection of rare (but aggressive) actions is a common phenomenon: Using the training samples, we compute the frequency with which the RL action was chosen by clinicians in each state. Among the 100 states where this proportion was smallest, the RL action is observed 1.5\% of the time on average (6.4 observations per state), while the action most frequently chosen by clinicians (\enquote{common practice}) is observed 35.7\% of the time on average (154.5 observations per state). These states collectively make up 26.2\% of the transitions in the training data, a nontrivial fraction. The RL policy tends to recommend more vasopressor treatment than common practice. For 99 of these 100 states, common practice involves zero vasopressors. Yet, the RL policy recommends vasopressors in 87 of those states, with 49 of those recommendations being large doses, which we define as those in the upper 50th percentile of nonzero amounts. Recent work has proposed various methods to constrain the RL policy to more closely resemble the behavior policy, which may be required here\cite{Futoma2020,fujimoto2019off}.

\textbf{Clinically implausible discharges may be related to discretization of time and censoring}: We observed for patient 1 that the model expects discharge even while on vasopressors.  This is clinically surprising but not obvious to the model given the data pre-processing: Of the 8271 training trajectories that end with discharge leading to survival, 4.9\% end on non-zero vasopressor dosages, and 2.6\% have large dosages, as defined in the previous paragraph. This may be because the data is discretized into 4-hour time intervals, an issue noted by other authors\cite{Jeter2019}, and the action captures the maximum vasopressor dosage during that interval. Thus, it is possible that vasopressors were administered briefly towards the start of that interval, and the patient was stabilized afterwards. As a result, the model contradicts clinical intuition by optimizing for discharge with vasopressors as a desirable goal:  Of the model-based roll-outs (based on test trajectories) that end with discharge leading to survival, 52.8\% end on non-zero vasopressor dosages, and 31.2\% end with large dosages. This may be driven in part by censoring: Of the 1714 training trajectories that get censored but eventually lead to 90-day survival, 10.3\% end on nonzero vasopressor dosages, and 5.6\% are large dosages. Even when we consider the censored samples, discharging a patient on vasopressors is still much more common in the roll-outs than would be expected in actual clinical practice.

\textbf{Other factors to investigate}:
We observed that the model may have insufficient knowledge about underlying conditions, including the surgical context for admission for patient 1 and specific comorbidities like lung cancer for patient 2. Features like these that could impact both treatment decisions and patient outcome may need to be added to the model to address unobserved confounding. The reward function is another part of the model that may be misspecified. For patient 2, the clinician could have already been aware during the ICU stay that patient comfort is the goal, while the RL model is still optimizing for patient survival. Learning better reward functions is an active area of research\cite{Srinivasan2020}. Our approach can help check whether the learned reward functions lead to models that align with expected outcomes.

%% file: sections/7-discussion.tex
\section*{Discussion} 
Reinforcement learning has the potential to support clinicians in patient care. To ensure that new policies are safe before deployment\cite{Futoma2020} and this exciting potential can reach fruition, we develop a strategy that allows researchers to bring clinicians into the study design process, identify potentially dangerous flaws in the model, and incorporate that information into models that are safe for deployment. To summarize, our workflow consists of three steps: (1) selecting trajectories for inspection based on surprisingly aggressive treatment recommendations or surprisingly positive outcomes, (2) inspecting the factual and model-based trajectories alongside the medical record, and (3) formulating insights for study design improvement. We demonstrate that applying our approach to a recent work on RL for sepsis suggests better reward specification, data preprocessing, and sample size considerations can help improve the design of the RL study. Our method only requires simulating from a learned model and can therefore be used in any model-based RL application, including more complex models involving neural networks\cite{Raghu2017} and latent variables\cite{futoma2020popcorn}. Our approach can also help build trust for models that accurately reflect patient dynamics, by allowing clinicians to check that the model correctly anticipates the impact of the policy it recommends. We see our workflow as a helpful procedure for checking model behavior in various applications of RL to healthcare, including ventilation weaning\cite{prasad2017reinforcement} and HIV\cite{parbhoo2017combining}, and a key step for bringing RL into meaningful clinical decision support tools.